\documentclass[10pt,twocolumn,letterpaper]{article}

\usepackage[pagenumbers]{cvpr} 

\usepackage[dvipsnames]{xcolor}
\usepackage[utf8]{inputenc} 
\usepackage[T1]{fontenc}    
\usepackage{url}            
\usepackage{booktabs}       
\usepackage{amsfonts}       
\usepackage{amsmath}
\usepackage{amsthm}
\usepackage{nicefrac}       
\usepackage{microtype}      
\usepackage{graphicx}
\usepackage{graphbox}
\usepackage{subcaption}
\usepackage{algorithm}
\usepackage{algorithmic}
\usepackage{bm}
\usepackage[accsupp]{axessibility} 
\usepackage{ulem}
\usepackage{colortbl,array}
\usepackage{multirow}
\usepackage{makecell}

\usepackage{pifont}
\usepackage{rotating}

\newcommand{\TODO}[1]{}
\renewcommand{\TODO}[1]{{\color{cyan} [TODO: {#1}]}}
\newcommand{\WIP}[1]{}
\renewcommand{\WIP}[1]{{\color{magenta} [WIP: {#1}]}}
\definecolor{midblue}{rgb}{0,0.11372549,0.258823529}
\newcommand{\argmax}{\mathop{\rm arg~max}\limits}

\newcommand{\inc}[1]{{\small (\textcolor{blue}{#1})}}
\newcommand{\dec}[1]{{\small (\textcolor{red}{#1})}}

\definecolor{cvprblue}{rgb}{0.21,0.49,0.74}
\usepackage[pagebackref,breaklinks,colorlinks,allcolors=cvprblue]{hyperref}

\def\paperID{12826} 
\def\confName{CVPR}
\def\confYear{2026}

\newtheorem{theorem}{Theorem}[section]
\definecolor{LightBlue}{rgb}{0.88,0.92,0.95} 
\definecolor{Blue}{rgb}{0.84,0.84,0.95} 
\definecolor{DeepBlue}{rgb}{0.66,0.66,0.95} 
\definecolor{LightYellow}{rgb}{0.98,0.96,0.89} 
\definecolor{LightRed}{rgb}{0.99,0.91,0.91}    

\title{Rationale-Enhanced Decoding for Multi-modal Chain-of-Thought}

\author{Shin'ya Yamaguchi\\
NTT\\
\and
Kosuke Nishida\\
NTT\\
\and
Daiki Chijiwa\\
NTT
}

\begin{document}
\maketitle

\begin{abstract}
Large vision-language models (LVLMs) have demonstrated remarkable capabilities by integrating pre-trained vision encoders with large language models (LLMs).
Similar to single-modal LLMs, chain-of-thought (CoT) prompting has been adapted for LVLMs to enhance multi-modal reasoning by generating intermediate rationales based on visual and textual inputs. 
While CoT is assumed to improve grounding and accuracy in LVLMs, our experiments reveal a key challenge: existing LVLMs often ignore the contents of generated rationales in CoT reasoning.
To address this, we re-formulate multi-modal CoT reasoning as a KL-constrained reward maximization focused on rationale-conditional log-likelihood.
As the optimal solution, we propose rationale-enhanced decoding (RED), a novel plug-and-play inference-time decoding strategy.
RED harmonizes visual and rationale information by multiplying distinct image-conditional and rationale-conditional next token distributions.
Extensive experiments show that RED consistently and significantly improves reasoning over standard CoT and other decoding methods across multiple benchmarks and LVLMs.
Our work offers a practical and effective approach to improve both the faithfulness and accuracy of CoT reasoning in LVLMs, paving the way for more reliable rationale-grounded multi-modal systems.
\end{abstract}

\begin{figure}[t]
    \centering
    \includegraphics[width=0.9\linewidth]{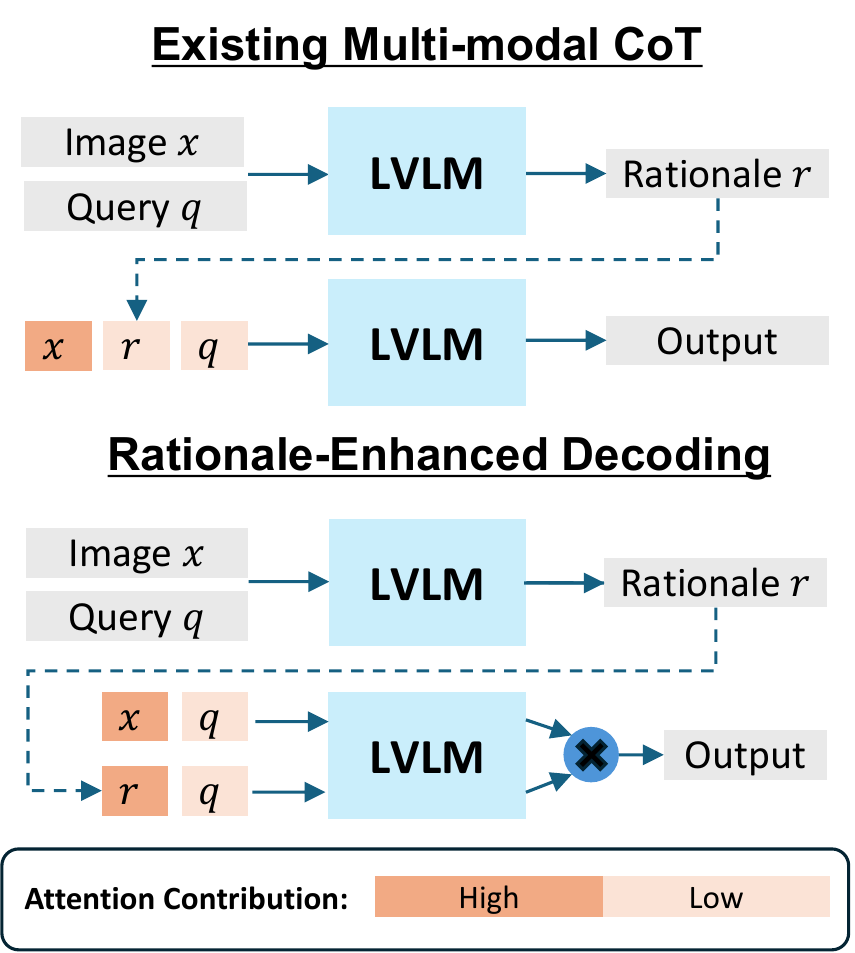}
    \caption{
    \textbf{Rationale-Enhanced Decoding (RED).}
    Existing multi-modal chain-of-thought (CoT) prompting by large vision language models (LVLMs) is a two-step generation of rationale and final output. It often focuses on input images and overlooks intermediate rationales in the final output generation.
    Our rationale-enhanced decoding (RED) addresses this issue by decoupling the image and rationale in decoding, and combining them at the logit level to provably ensure grounding outputs on the rationale.
    }
    \label{fig:top}
    \vspace{-5mm}
\end{figure}

\section{Introduction}
Recent large language model (LLM) advancements exhibit impressive complex reasoning capabilities~\cite{Achiam_2023_GPT4,Google_2023_Gemini,Bai_2022_Claude}. 
This progress now extends beyond single-modal text-to-text reasoning.
The integration of pre-trained vision encoders, like CLIP~\cite{Radford_ICML21_CLIP}, with LLMs has led to the development of large vision-language models (LVLMs)~\cite{liu_NeurIPS23_llava,Dai_NeurIPS23_InstructBLIP,Zhu_ICLR24_MiniGPT,Chen_NeurIPS24_InternVL,Bai_2025_Qwen-vl}, enabling more complex multi-modal reasoning.

Chain-of-thought (CoT) prompting is a key factor in LLM reasoning abilities~\cite{Wei_NeurIPS22_CoT,Kojima_NeurIPS22_lets_think_step_by_step}. 
CoT-prompted models first generate intermediate reasoning steps, termed \textit{rationales}, then incorporate them into the context to produce the final query response.
CoT facilitates models to understand queries deeply and promotes logical and coherent responses~\cite{Wang_ACL23_towards_understanding_cot,Saparov_ICLR23_systematic_analysis_cot,Wang_COLING24_analyzing_cot}.
This principle has also been applied to multi-modal reasoning in LVLMs~\cite{Zhang_TMLR_mm-cot,Lu_NeurIPS22_unifiedqa,Zheng_NeurIPS23_ddcot,He_AAAI24_mm_latent_cot,Mondal_AAAI24_kam-cot,Mitra_CVPR24_CCoT,Zheng_LREC24_enhancing_mmcot_soft_negative_sampling,Zheng_ACMMM24_mm_debate}. 
In CoT for LVLMs, models generate intermediate rationales from the input image and the text query. 
These rationales, with the original inputs, are then used for grounded image-text-to-text reasoning.
Recent research also explores generating structured rationales (e.g., scene graphs) to enhance LVLMs' capabilities like spatial reasoning~\cite{Mitra_CVPR24_CCoT,Zheng_ACMMM24_mm_debate}.
Thus, CoT in LVLMs is widely assumed to be beneficial, offering multi-modally grounded understanding for more accurate responses.

However, our empirical results show that this assumption does not always hold.
Our preliminary experiments, including (i) measuring input token contributions and (ii) swapping rationales, reveal a critical CoT limitation: existing LVLMs often ignore the contents of generated rationales.
Indeed, CoT reasoning with rationales can even degrade performance compared to direct answering without CoT.
Moreover, replacing a rationale with an irrelevant one often does not change model performance, implying LVLMs largely ignore rationale semantics in such cases.
These findings suggest that the current CoT mechanism in LVLMs does not effectively ground the final prediction on the information captured by the intermediate rationale.

A straightforward solution to make models force rationale grounding would be to fine-tune LVLMs on datasets containing image-query-rationale-answer tuples, as in~\cite{Zhang_TMLR_mm-cot,Zheng_NeurIPS23_ddcot}.
However, this approach demands expensive annotated datasets and additional training costs.
Therefore, we focus on developing a plug-and-play decoding strategy for pre-trained LVLMs. 
We address the following primary research question: \textit{Can we enhance the LVLMs' grounding capability on the rationales in CoT and improve performance by solely modifying the decoding strategy without additional training?}

In this paper, we propose rationale-enhanced decoding (RED), a novel decoding strategy designed to harmonize information from visual and rationale tokens in LVLMs without additional training.
Our core idea is decoupling the next token probability into distinct image-conditional $p(y|x,q)$ and rationale-conditional $p(y|r,q)$ and separately enhancing rationale grounding ($x$: image, $r$: rationale, $q$: query, and $y$: output).
Based on this idea, we re-formulate multi-modal CoT prompting as a KL-constrained reward maximization~\cite{Schulman_ICML15_TRPO,Schulman_ICML15_PPO}.
This aims to maximize policy regarding the rationale-conditional log-likelihood $\log p(y|r,q)$ as the reward while staying close to the image-conditional likelihood $p(y|x,q)$.
Solving this problem makes LVLMs explicitly ground on both image and rationale information in the next token prediction.
Without additional training, RED provably yields the optimal solution of this maximization by composing the next-token distribution as $p(y|x,q)\times p(y|r,q)^\lambda$.
Practically, RED is implemented as a simple weighted sum of the log-softmax logits of $p(y|x,q)$ and $p(y|r,q)$, allowing easy, training-free, and plug-and-play integration with existing LVLMs without architecture modification.

We conduct extensive experiments comparing RED against standard CoT prompting and plug-and-play decoding methods for off-the-shelf LVLMs across multiple benchmark datasets and backbone LVLMs.
Our results demonstrate that RED consistently and significantly improves reasoning performance.
Furthermore, RED's advantages are amplified with high-quality rationales (e.g., intervening with GPT-4\cite{Achiam_2023_GPT4}).
Our findings validate the effectiveness and practicality of RED for enhancing CoT reasoning faithfulness and accuracy in LVLMs, opening avenues for more reliable, interpretable, rationale-grounded multi-modal systems.

\section{Related Work}
\paragraph{Large Vision Language Models (LVLMs).}
Large vision language models (LVLMs) integrate visual encoders with LLMs, often via alignment training to represent images in the LLM input space~\cite{liu_NeurIPS23_llava, Dai_NeurIPS23_InstructBLIP,Zhu_ICLR24_MiniGPT,Chen_NeurIPS24_InternVL}.
Despite remarkable multi-modal capabilities~\cite{Laurenccon_NeurIPS24_matters_of_lvlms,Parekh_NeurIPS24_conceptual_explainability_for_lvlms,Neo_ICLR25_interpetting_lvlms}, LVLMs face challenges like poor visual recognition~\cite{Zhang_NeurIPS24_why_lvlm_bad_classification}, object hallucination~\cite{Li_EMNLP23_evaluating_lvlm,Gunjal_AAAI24_detecting_object_hallucination_lvlm,Wu_ICML24_analyzing_hallucination_lvlm}, and misalignment between image and text tokens~\cite{Zeng_CVPR24_investigating_compositionality_challenge_lvlm,Wang_NeurIPS24_picture_worth_lvlm,Campbell_NeurIPS24_understanding_binding_limits_lvlms}.
Furthermore, as our preliminary experiments (Section~\ref{sec:motivation}) highlight, LVLMs also struggle to effectively leverage rationales in CoT reasoning.
Previous solutions include preference/reward tuning~\cite{Zhou_NeurIPS24_calibrated_reward_tuning,Sun_ACL24F_aligning_lvlm_augmented_rlhf}, improved visual instruction tuning~\cite{Ossowski_ACL24_olive,Chen_ICLR25_perturbollava,Shi_ICLR25_enhancing_cognition_lvlm,Chow_ICLR24_unified_training_lvlm,Sirnam_ECCV24_x-former}, auxiliary model-enhanced decoding~\cite{Yu_ECCV24_attention_prompting,Zhang_ICLR25_selfcorrecting_generative_decoding_lvlm,Sun_ICLR25_fast_slow_visual_agents}, plug-and-play decoding~\cite{Manevich_ACL24_LCD,Leng_CVPR24_VCD,Wang_ACL24F_ICD,Ghosh_ICLR25_VDGD}, and CoT prompting~\cite{Zhang_TMLR_mm-cot,Zheng_NeurIPS23_ddcot,Mondal_AAAI24_kam-cot,He_AAAI24_mm_latent_cot,Zheng_LREC24_enhancing_mmcot_soft_negative_sampling,Mitra_CVPR24_CCoT}.
Our RED is categorized into a plug-and-play decoding for improving CoT prompting.
While many CoT prompting methods focus on improving rationale generation via specific prompting or auxiliary model training (see Section~\ref{sec:mm-cot}), they often still rely on standard decoding with $p_\theta(y_i|\bm{y}_{<i},x,r,q)$ that may not leverage rationales. 
Our work diverges with a novel plug-and-play decoding strategy designed to enhance rationale utilization in CoT reasoning, without additional training or auxiliary models.

\paragraph{Plug-and-play Decoding Strategies for LVLMs.}
Plug-and-play decoding strategies~\cite{Manevich_ACL24_LCD, Leng_CVPR24_VCD,Wang_ACL24F_ICD, Kim_NeurIPS24_code, Ghosh_ICLR25_VDGD} are relevant as they operate at inference time without additional training like RED.
These decoding strategies are mainly focused on mitigating object hallucination by contrastive decoding~\cite{Li_ACL23_contrastive_decoding}.
For instance, LCD~\cite{Manevich_ACL24_LCD} contrasts $p_\theta(y_i|\bm{y}_{<i},x,q)$ with $p_\theta(y_i|\bm{y}_{<i},q)$ to mitigate the language prior effects; VCD~\cite{Leng_CVPR24_VCD} subtracts hallucinated predictions by contrasting $p_\theta(y_i|\bm{y}_{<i},x,q)$ with $p_\theta(y_i|\bm{y}_{<i},x^\prime,q)$ ($x^\prime$ is the corrupted image).
Thus, these decoding strategies basically aim to modulate image-conditional probability $p_\theta(y_i|\bm{y}_{<i},x,q)$ to mitigate object hallucination.
However, these methods do not ensure LVLMs faithfully use CoT rationales; they refine image-conditional output, not harmonize it for CoT reasoning.
Therefore, RED is complementary to these decoding strategies because it grounds predictions by multiplying distinct image-conditional $p_\theta(y_i|\bm{y}_{<i},x,q)$, which is potentially pre-modulated by other methods, and rationale-conditional $p_\theta(y_i|\bm{y}_{<i},r,q)$.
This allows combining RED with other plug-and-play methods for synergistic benefits of better hallucination mitigation and rationale grounding.

\section{Preliminaries}
This section introduces LVLM and CoT prompting principles, then presents preliminary experiments highlighting existing LVLMs' suboptimal rationale utilization in multi-modal CoT.

\subsection{Next-token Prediction in LVLMs}
Consider an auto-regressive LVLM, parameterized by $\theta$, which is trained on large-scale image-text datasets to process images as input for its backbone LLM.
Given an input image $x$ and query $q$, an LVLM generates an output token sequence $\bm{y}=(y_1,\dots,y_L)\in\mathcal{V}^L$ following:
\begin{equation}
     p(\bm{y}|x,q) = \prod^{L}_{i=1}p_\theta(y_i|\bm{y}_{<i},x,q),\label{eq:lvlm_cond_dist}
\end{equation}
where $L$ is token length, $\mathcal{V}$ is token vocabulary, and $\bm{y}_{<i}$ are preceding output tokens.
As in auto-regressive LLMs, $p_\theta(y_i|\bm{y}_{<i},x,q)$ over $\mathcal{V}$ is the softmax of the model's output $\operatorname{logits}_\theta(y_i|\bm{y}_{<i},x,q)$:
\begin{gather}
     p_\theta(y_i|\bm{y}_{<i},x,q) = \operatorname{softmax}(\operatorname{logits}_\theta(y_i|\bm{y}_{<i},x,q)) \\
     = \frac{\exp(\operatorname{logits}_\theta(y_i|\bm{y}_{<i},x,q))}{\sum_{{w}\in\mathcal{V}}\exp(\operatorname{logits}_\theta(y_i=w|\bm{y}_{<i},x,q))}.
\end{gather}
Each token $y_i$ is generated from $p_\theta(y_i|\bm{y}_{<i},x,q)$ via a decoding strategy such as greedy decoding, i.e., $y_i = \argmax_{w\in\mathcal{V}}p_\theta(y_i|\bm{y}_{<i},x,q)$.
We define $\operatorname{generate(\cdot)}$ as a utility function for an arbitrary decoding strategy:
\begin{equation}
     \bm{y} = \operatorname{generate}_\theta(x,q).\label{eq:lvlm_generate}
\end{equation}

\subsection{Multi-modal Chain-of-Thought Prompting}\label{sec:mm-cot}
Inspired by single modal CoT prompting~\cite{Wei_NeurIPS22_CoT}, multi-modal CoT aims to enhance LVLM reasoning by incorporating the intermediate rationale generation during decoding~\cite{Zhang_TMLR_mm-cot,Zheng_NeurIPS23_ddcot,Mondal_AAAI24_kam-cot,Mitra_CVPR24_CCoT}.
In general, multi-modal CoT involves two reasoning steps: (i) rationale generation and (ii) output generation.
First, LVLMs generate a rationale $r$ from input image $x$, query prompt $q$ with instruction prompt, e.g., ``\texttt{Given the image, generate the rationale for answering the question}'', as follows.
\begin{equation}
     r = \operatorname{generate}_\theta(x,q).\label{eq:rationale_generation}
\end{equation}
By using $r$, LVLMs generate the final output by the next-token prediction, similar to Eq.~(\ref{eq:lvlm_generate}) :
\begin{align}
     \bm{y} = \operatorname{generate}_\theta(x,r,q).\label{eq:lvlm_cot_generate}
\end{align}
In this line of work, MM-CoT~\cite{Zhang_TMLR_mm-cot} and UnifiedQA~\cite{Lu_NeurIPS22_unifiedqa} are the pioneering works introducing the concept of CoT prompting into the multi-modal reasoning of LVLMs.
Successor works focus on improving the quality of $r$ via auxiliary VQA models~\cite{Zheng_NeurIPS23_ddcot} and knowledge base retrieval~\cite{Mondal_AAAI24_kam-cot}.
Although these works successfully elicited LVLM reasoning, their reliance on additional training and/or auxiliary resources (e.g., VQA models and knowledge bases) limits broader applicability.
To address this limitation, CCoT~\cite{Mitra_CVPR24_CCoT} enhanced rationale quality via structured scene graphs in JSON format generated from LVLMs' zero-shot reasoning without additional training.
Most existing works focus on improving $r$ in Eq.~(\ref{eq:rationale_generation}) and assume that accurate $r$ ensures better next token prediction with $p_\theta(y_i|\bm{y}_{<i},x,r,q)$.
In this regard, our work is orthogonal to them because we investigate the reliability of $p_\theta(y_i|\bm{y}_{<i},x,r,q)$ as discussed in the next section.
This work focuses on training-free CoT reasoning for maintaining simplicity and generalization of pre-trained LVLMs, but our findings can extend to any CoT methods in a plug-and-play manner.\looseness-1

\begin{figure*}[t] 
  \centering
  \begin{minipage}[]{0.48\linewidth}
    \centering
    \includegraphics[width=0.85\linewidth]{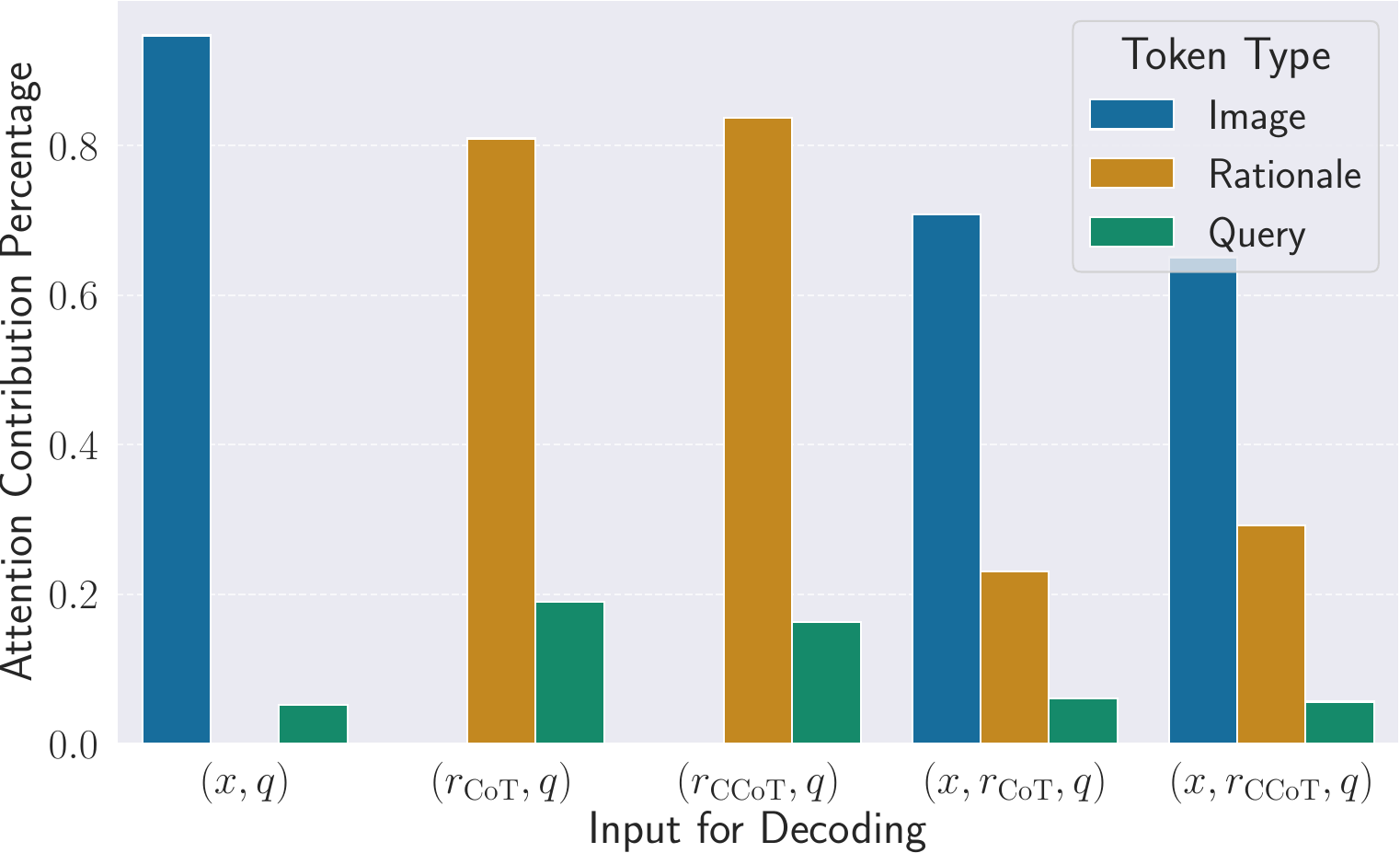}
    \captionof{figure}{\textbf{Percentage of attention contributions by input token types for different decoding strategies (Gemma-3-12B)}. Rationale tokens contribute less to outputs than image tokens.}
    \label{fig:preexp_attention_contribution}
  \end{minipage}
  \hspace{\fill}
  \begin{minipage}[]{0.48\linewidth}
    \centering
    \resizebox{0.85\textwidth}{!}{
    \begin{tabular}{lccc}
      \toprule
      Input for Decoding & Gemma-3-4B & Gemma-3-12B \\
      \midrule
      $(x,q)$ & 40.00 & 45.34 \\\midrule
      $(x, r_\mathrm{CoT}, q)$ & 41.08 \inc{+1.08} & 41.76 \dec{-3.58} \\
      $(x, r^\prime_\mathrm{CoT}, q)$ & 41.88 \inc{+1.88} & 41.75 \dec{-3.59} \\
      $(x, r_\mathrm{CCoT}, q)$ & 44.54 \inc{+4.54} & 44.50 \dec{-0.84} \\
      $(x, r^\prime_\mathrm{CCoT}, q)$ & 44.35 \inc{+4.35} & 44.30 \dec{-1.04} \\\midrule
      $(r_\mathrm{CoT}, q)$ & 40.15 \inc{+0.15} & 37.87 \dec{-7.47} \\
      $(r^\prime_\mathrm{CoT}, q)$ & 7.40 \dec{-32.60} & 16.21 \dec{-29.13}\\
      $(r_\mathrm{CCoT}, q)$ & 43.35 \inc{+3.35} & 44.23 \dec{-1.11} \\
      $(r^\prime_\mathrm{CCoT}, q)$ & 19.05 \dec{-20.95} & 10.25 \dec{-35.09} \\
      \bottomrule
    \end{tabular}
    }
    \captionof{table}{\textbf{Test accuracy (\%) on GQA}. Contrary to intuition, irrelevant rationale $r^\prime$ maintains the model performance in decoding with $(x,r^\prime,q)$, indicating the ignorance of rationales in LVLMs.}
    \label{tab:preexp_intervention}
  \end{minipage}
  \vspace{-5mm}
\end{figure*}

\subsection{Motivating Experiments}\label{sec:motivation}
We show our motivation through preliminary experiments asking a simple question: \textit{Do LVLMs perform CoT reasoning grounded on intermediate rationales?}
Specifically, we assess how much $r$ and its contents contribute to the output sequence decoded by $p_\theta(y_i|\bm{y}_{<i},x,r,q)$ through two preliminary experiments: (i) measuring the contribution scores to output token predictions for input token groups corresponding to the image $x$, rationale $r$, and query $q$, and (ii) replacing rationale $r$ with $r^{\prime}$ from another $(x^\prime,q^\prime)$ pair.
Experiment (i) aims to evaluate how much $r$ contributes to output $\bm{y}$, and (ii) checks if LVLMs leverage the content of $r$ associated with $(x,q)$.
We used Gemma-3-4B/12B~\cite{Google_2025_gemma3} as the LVLM and GQA~\cite{Hudson_CVPR19_gqa} as the evaluation dataset.
For the rationale generation, we queried LVLMs with Eq.~(\ref{eq:rationale_generation}) to generate text description rationales and scene graphs with the prompts of CCoT~\cite{Mitra_CVPR24_CCoT}; hereinafter, we refer to the reasoning with text descriptions as CoT and that with scene graphs as CCoT.
See Appendix for details.

\noindent\textbf{Weakened influence of rationales.}
Here, we measure attention contribution scores~\cite{Kobayashi_EMNLP20_attention_contribution,Basu_NeurIPS24_understanding_info_lvlm,Kang_ICLR25_visual_attention_sink} to quantify input token $(x,r,q)$ contributions to output $\bm{y}$.
Ideally, CoT prediction should derive substantial contributions from both image and rationale tokens.
The contribution score from $i$-th token to $j$-th token at the $l$-th layer and $h$-th head is computed by $\|\alpha^{l,h}_{i,j}\bm{z}^{l-1}_j\bm{W}^{l,h}_{OV}\|$, where $\alpha^{l,h}_{i,j}$ is attention score, $\bm{z}^{l-1}_j$ is the output of the previous layer, and $\bm{W}^{l,h}_{OV}$ is the output projection matrix in transformer-based LVLMs.
We describe more details of the evaluation protocol in Appendix. 
Figure~\ref{fig:preexp_attention_contribution} shows the attention contributions in the middle layer of Gemma-3-12B for each case of input for decoding, where the scores are computed for each token corresponding to $x$, $r$, and $q$, respectively, and displayed as percentages of the overall contribution score for each input type \footnote{We analyzed a middle layer because prior work shows it is where LVLMs primarily perform multi-modal fusion to integrate vision and language~\cite{Neo_ICLR25_interpetting_lvlms}. This allows us to observe the model’s attention balance between the image and rationale during this crucial integration process, before the representation becomes overly specialized for the final output task.}.
While image and rationales contribute largely when conditioned individually (i.e., $(x,q)$, $(r_\mathrm{CoT},q)$, and $(r_\mathrm{CCoT},q)$), in multi-modal CoT (i.e., $(x,r_\mathrm{CoT},q)$, and $(x,r_\mathrm{CCoT},q)$), image tokens dominate and rationale contribution is remarkably reduced.

\noindent\textbf{Lack of grounding on rationale contents.}
Next, we analyze rationale content effects by replacing original rationale $r$ with an irrelevant rationale $r^\prime$ generated from another random $(x^\prime,q^\prime)$ pair.
We expect that intervening by $r^\prime$ will largely degrade the performance.
Table~\ref{tab:preexp_intervention} shows the test accuracy on GQA for the combinations of decoding strategy and input for decoding.
While CoT and CCoT with $(x,r,q)$ improved the baseline performance with Gemma-3-4B, they degraded the performance with Gemma-3-12B.
More importantly, in Table~\ref{tab:preexp_intervention}, CoT/CCoT with the irrelevant rationale $r^\prime$ via $(x,r^\prime,q)$ performed similarity to using original $r$, indicating that LVLMs ignore the contents of $r$ and the performance gain comes from other causes.
Conversely, decoding with $(r^\prime,q)$ largely dropped the scores from that with $(r,q)$, implying that LVLMs use $r$ when $x$ is absent.
This phenomenon could potentially be attributed to several factors, including position bias in the LLM backbones~\cite{Zhang_NeurIPS24_pos_bias,Zhang_EMNLP24_instruct_pos_bias}, attention sink~\cite{Xiao_ICLR24_attention_sink,Gu_ICLR25_when_attention_sink_emerges,Kang_ICLR25_visual_attention_sink}, and/or overfitting to visual instruction tuning~\cite{Li_EMNLP23_evaluating_lvlm,He_ICLR25_analyzing_FGVR_lvlm}.
Since our focus is training-free CoT in off-the-shelf LVLMs, we leave a deeper analysis for future work.\looseness-1

In summary, the two experiments demonstrate that LVLMs are less likely to leverage rationale $r$ mechanistically and semantically in the multi-modal CoT prompting.
This challenge is important because CoT prompting is generally expected to enhance not only task performance but also faithfulness and interpretability~\cite{Lanham_2023_measuring_faithfulness_cot,Paul_EMNLP24_measuring_improving_faithfulness_cot,Jie_NAACL24_interpretability_cot}.
This motivates our new decoding strategy to ground the inference on $r$.\looseness-1

\section{Method}
To address the previously identified challenges, we propose rationale-enhanced decoding (RED).
Our approach begins by re-formulating multi-modal CoT prompting as a KL-constrained reward maximization focused on rationale-conditional probability.
RED solves this without additional training by multiplying distinct image- and rationale-conditional token probabilities for predicting the next token.
We prove that this formulation is equivalent to the optimal solution of the KL-constrained reward maximization, demonstrating its theoretically sound background.
RED is implemented by a weighted sum of the log-softmax logits, allowing for its plug-and-play adaptation to any LVLMs.

\subsection{Multi-modal CoT as a Reward Maximization on Rationale}
As shown in Section~\ref{sec:motivation}, conventional multi-modal CoT prompting via $p_\theta(y_i|\bm{y}_{<i},x,r,q)$ often fails to effectively leverage rationale $r$ for output predictions.
We also observed that the prediction with $p_\theta(y_i|\bm{y}_{<i},r,q)$ relies on $r$, but the performance is suboptimal.
We aim for a next token distribution more effective than both $p(y_i|\bm{y}_{<i},x,r,q)$ and $p(y_i|\bm{y}_{<i},r,q)$.
To this end, our core idea is decoupling the next token probability into distinct image-conditional $p_\theta(y_i|\bm{y}_{<i},x,q)$ and rationale-conditional $p_\theta(y_i|\bm{y}_{<i},r,q)$ and separately enhancing rationale grounding.
Enhancing rationale grounding indeed corresponds to maximizing the log-likelihood of $p_\theta(y_i|\bm{y}_{<i},r,q)$; we refer to this log-likelihood as \textit{rationale-grounding reward}.
Therefore, instead of $p_\theta(y_i|\bm{y}_{<i},x,r,q)$, we introduce a new next token distribution $\pi$ maximizing the rationale-grounding reward based on a KL-constrained reward maximization~\cite{Schulman_ICML15_TRPO,Schulman_ICML15_PPO} as follows:
\begin{equation}
    \max_{\pi} \mathbb{E}_{\pi}[R] - \beta \mathbb{D}_\mathrm{KL}[\pi || \pi_\mathrm{ref}],\label{eq:red_objective}
\end{equation}
where $R=\log p_\theta(y_i \sim \pi|\bm{y}_{<i},r,q)$, $\beta$ is a hyperparameter to balance the KL penalty term, and $\pi_\mathrm{ref}=p_\theta(y_i|\bm{y}_{<i},x,q)$.
Intuitively, this maximizes expected rationale-grounding reward $\mathbb{E}_{y_i\sim \pi}[\log p_\theta(y_i|\bm{y}_{<i},r,q)]$, which strongly relies on $r$ as shown in Section~\ref{sec:motivation}.
The KL-constraint between $\pi$ and $p(y_i|\bm{y}_{<i},x,q)$ incorporates visual information from $x$.
This formulation naturally makes the next token prediction ground on both $x$ and $r$ without using the problematic conditional probability $p_\theta(y_i|\bm{y}_{<i},x,r,q)$.

\subsection{Rationale-Enhanced Decoding (RED)}\label{sec:red}
We derive rationale-enhanced decoding (RED) to maximize Eq.~(\ref{eq:red_objective}) without additional training.
RED's next token distribution $\hat{p}_\theta(y_i)$ is formed by multiplying distinct distributions as follows:
\begin{equation}
     \hat{p}_\theta(y_i) := \frac{1}{Z_\theta} p_\theta(y_i|\bm{y}_{<i},x,q) \times p_\theta(y_i|\bm{y}_{<i},r,q)^\lambda,\label{eq:red}
\end{equation}
where $Z_\theta=\sum_{w\in\mathcal{V}} p_\theta(y_i=w|\bm{y}_{<i},x,q) \times p_\theta(y_i=w|\bm{y}_{<i},r,q)^\lambda$ is the normalization constant, and $\lambda$ is a hyperparameter for modulating the influence of $p_\theta(y_i|\bm{y}_{<i},r,q)$.
Intuitively, Eq.~(\ref{eq:red}) is a power-of-experts~\cite{Hinton2002_poe} emphasizing overlap between image- and rationale conditional probabilities.
Indeed, Eq.~(\ref{eq:red}) is the closed-form solution maximizing Eq.~(\ref{eq:red_objective}) as follows.
\begin{theorem}\label{prop:red}
Let the reference policy $\pi_\mathrm{ref}$ be $p_\theta(y_i|\bm{y}_{<i},x,q)$, and the reward function $R(\cdot)$ be $\log p_\theta(y_i|\bm{y}_{<i}, r, q)$.
Sampling by Eq.~(\ref{eq:red}) is equivalent to sampling from the optimal policy $\pi^*$ for Eq.~(\ref{eq:red_objective}).
\end{theorem}

\begin{proof}
Consider the KL-constrained reward maximization objective~\cite{Schulman_ICML15_TRPO,Schulman_ICML15_PPO,Rafailov_NeurIPS23_DPO}:
\begin{equation}
    \max_{\pi} \mathbb{E}_{\pi}[R(s,a)] - \beta \mathbb{D}_\mathrm{KL}[\pi(a|s) || \pi_\mathrm{ref}(a|s)],\label{eq:klmax_problem}
\end{equation}
where $R$ is a reward function, $s$ is a state (input context), and $a$ is an action (output).
Obviously, Eq.~(\ref{eq:red_objective}) is a special case of Eq.~(\ref{eq:klmax_problem}).
From Appendix A.1 of \cite{Rafailov_NeurIPS23_DPO}, the optimal policy $\pi^*(a|s)$ for this objective is given by
\begin{equation}
    \pi^*(a|s) = \frac{1}{Z(s)} \pi_\mathrm{ref}(a|s) \exp\left(\frac{1}{\beta} R(s,a)\right),\label{eq:klmax_optimal}
\end{equation}
where $Z(s) = \sum_{a^\prime} \pi_\mathrm{ref}(a^\prime|s) \exp(\frac{1}{\beta} R(s,a^\prime))$ is the partition function.
In the maximization of Eq.~(\ref{eq:red_objective}), given $a=y_i$ and $s=(\bm{y}_{i<},x, r,q)$, we can set each component in Eq.~(\ref{eq:klmax_problem}) as $\pi_\mathrm{ref}(a|s) = p_\theta(y_i|\bm{y}_{<i},x,q)$, $R(s,a)=\log p_\theta(y_i|\bm{y}_{<i},r,q)$, and $\beta = 1/\lambda$.
Substituting them into the optimal policy of Eq.~(\ref{eq:klmax_optimal}) yields
\begin{align*}
\pi^*(a|s) &= \frac{1}{Z_\theta}p_\theta(y_i|\bm{y}_{<i},x,q) \times \exp\left(\lambda \log p_\theta(y_i|\bm{y}_{<i},r,q)\right) \\
&= \frac{1}{Z_\theta}p_\theta(y_i|\bm{y}_{<i},x,q) \times p_\theta(y_i|\bm{y}_{<i},r,q)^\lambda = \text{Eq.~(\ref{eq:red})}.
\end{align*}
Therefore, sampling by Eq.~(\ref{eq:red}) is indeed equivalent to sampling from the optimal policy $\pi^*(a|s)$ for the KL-constrained reward maximization in Eq.~(\ref{eq:red_objective}).
\end{proof}

Theorem~\ref{prop:red} shows that sampling by Eq.~(\ref{eq:red}) yields the optimal distribution maximizing $p_\theta(y_i|\bm{y}_{<i},r,q)$ at decoding time without any auxiliary reward models.
This theoretical property supports the reliability of RED as a method for improving LVLMs in the multi-modal CoT.
We also show a comparison to other possible alternatives to validate the efficacy of Eq.~(\ref{eq:red}) in Appendix.

\subsection{Algorithm}
Practically, we generate the next token by combining $\operatorname{logits}_\theta(y_i|\bm{y}_{<i},x,q)$ and $\operatorname{logits}_\theta(y_i|\bm{y}_{<i},r,q)$:
\begin{gather}
     \hat{p}_\theta(y_i) = \operatorname{softmax}(\hat{\operatorname{logits}}_\theta(y_i)), \label{eq:red_prob}\\
     \hat{\operatorname{logits}}_\theta(y_i) := \log\operatorname{softmax}(\operatorname{logits}_\theta(y_i|\bm{y}_{<i},x,q)) \nonumber\\
     + \lambda \log\operatorname{softmax}(\operatorname{logits}_\theta(y_i|\bm{y}_{<i},r,q)).\label{eq:red_logits}
\end{gather}
Eqs.~(\ref{eq:red_prob})~and~(\ref{eq:red_logits}) are derived from Eq.~(\ref{eq:red}) by
\begin{align}
    &\hat{p}_\theta(y_i) = \exp\log(\hat{p}_\theta(y_i)) \nonumber\\
    &= \exp(\log(p_\theta(y_i|\bm{y}_{<i},x,q))\!+\!\log( p_\theta(y_i|\bm{y}_{<i},r,q)^\lambda)\!-\!\log Z_\theta)\nonumber\\
    &\propto\! \operatorname{softmax}(\log(p_\theta(y_i|\bm{y}_{<i},\!x,\!q)\!\times\! p_\theta(y_i|\bm{y}_{<i},\!r,\!q)^\lambda)).
\end{align}
Thus, we can implement RED by computing the weighted sum of the log-softmax logits as the new logits for the next token prediction.
To avoid the latency overhead, we simultaneously compute $\operatorname{logits}_\theta(y|\bm{y},x,q)$ and $\operatorname{logits}_\theta(y|\bm{y},r,q)$ by batch parallel inference.
We show the overall procedures of RED in Algorithm~\ref{alg:red}.

\begin{figure}
\vspace{-2mm}
\begin{algorithm}[H]
    \caption{Rationale-Enhanced Decoding (RED)}\label{alg:red}
    \begin{algorithmic}[1]
        \REQUIRE{Input image $x$, query $q$, LVLM parameterized by $\theta$, hyper-parameter $\lambda$}
        \ENSURE{Decoded output sequence $\bm{y}$}
        \STATE{$r \leftarrow \operatorname{generate}_\theta(x,q)$~~~~\text{\color{gray}\# Generate arbitrary rationales}}
        \STATE{$\bm{y} \leftarrow \emptyset$}
        \FOR{$|\bm{y}| < L$}
            \STATE{$\hat{\operatorname{logits}_\theta}(y) \leftarrow \log\operatorname{softmax}(\operatorname{logits}_\theta(y|\bm{y},x,q)) + \lambda \log\operatorname{softmax}(\operatorname{logits}_\theta(y|\bm{y},r,q))$}
            \STATE{$y\sim \operatorname{softmax}(\hat{\operatorname{logits}_\theta}(y))$}
            \STATE{$\bm{y}\operatorname{.append}(y)$~~~~\text{\color{gray}\# Add last token}}
        \ENDFOR
    \end{algorithmic}
\end{algorithm}
\vspace{-8mm}
\end{figure}
\section{Experiments}
We validate the efficacy of RED via comprehensive evaluation on multiple multi-modal reasoning datasets with pre-trained LVLMs, comparing to existing multi-modal CoT methods and plug-and-play decoding baselines.
We also analyze interventions on the rationale quality.

\begin{table*}[t]
\centering
\caption{\textbf{Performance comparison on general visual reasoning benchmarks across various LVLMs}. RED is applied to both CoT and CCoT. Values in parentheses indicate the relative delta from Baseline (direct decoding without CoT). Best scores for each LVLM-benchmark pair are bolded.
}
\label{tab:general_benchmark}
\resizebox{\linewidth}{!}{
\begin{tabular}{@{}lcccccccc}
\toprule
 & \multirow{2}{*}{\textbf{GQA}} & \multirow{2}{*}{\textbf{LLaVA-Bench}} & \multicolumn{2}{c}{\textbf{MME}} & \multirow{2}{*}{\textbf{MMVet}} & \multirow{2}{*}{\textbf{SEED-I}} & \multirow{2}{*}{\textbf{TextVQA}} & \multirow{2}{*}{\textbf{MathVista}} \\
\cmidrule(r){4-5}
 &  &  & \textbf{Perception} & \textbf{Cognition} &  \\
\midrule
\multicolumn{9}{@{}l}{\cellcolor{LightBlue}\textbf{Gemma-3-4B}} \\
Baseline & 40.00 & 73.20 & 1211.34 & 370.00 & 44.00 & 65.39 & 63.95 & 41.40 \\
VCD & 38.74 \dec{-1.26} & 75.70 \inc{+2.50} & 1184.11 \dec{-27.23} & 338.57 \dec{-31.43} & 44.80 \inc{+0.80} & 65.19 \dec{-0.20} & 64.02 \inc{+0.07} & 39.30 \dec{-2.10}\\
VCD + ICD & 38.62 \dec{-1.33} & 73.30 \inc{+0.10} & 1182.29 \dec{-29.05} & 340.71 \dec{-29.29} & 46.70 \inc{+2.70} & 65.20 \dec{-0.19} & 64.03 \inc{+0.08} & 40.40 \dec{-1.00}\\
CoT & 41.08 \inc{+1.08} & 72.90 \dec{-0.30} & 1254.77 \inc{+43.43} & 341.07 \dec{-28.93} & 46.20 \inc{+2.20} & 65.90 \inc{+0.51} & 60.62 \dec{-3.33} & 40.10 \dec{-1.30} \\
CCoT & 44.54 \inc{+4.54} & 73.60 \inc{+0.40} & 1294.46 \inc{+83.12} & 396.43 \inc{+26.43} & 46.50 \inc{+2.50} & 66.63 \inc{+1.24} & 63.23 \dec{-0.72} & 41.10 \dec{-0.30} \\
CoT + RED & 42.19 \inc{+2.19} & 76.30 \inc{+3.10} & 1325.77 \inc{+114.43} & \textbf{645.35 \inc{+275.35}} & 46.60 \inc{+2.60} & \textbf{66.89 \inc{+1.50}} & 65.13 \inc{+1.18} & \textbf{43.50 \inc{+2.10}} \\
CCoT + RED & \textbf{45.87 \inc{+5.87}} & \textbf{76.90 \inc{+3.70}} & \textbf{1330.07 \inc{+118.73}} & 611.43 \inc{+241.43} & \textbf{49.10 \inc{+5.10}} & 66.73 \inc{+1.34} & \textbf{65.54 \inc{+1.59}} & 42.00 \inc{+0.60}\\
\midrule
\multicolumn{9}{@{}l}{\cellcolor{blue!25}\textbf{Gemma-3-12B}} \\ 
Baseline & 45.34 & 79.00 & 1171.37 & 545.71 & 57.70 & 71.01 & 69.81 & 52.10 \\
VCD & 44.11 \dec{-1.23} & 78.40 \dec{-0.60} & 1152.02 \dec{-19.34} & 543.93 \dec{-1.79} & 58.30 \inc{+0.60} & 70.96 \dec{-0.05} & 69.40 \dec{-0.51} & 51.50 \dec{-0.60}\\
VCD + ICD & 44.16 \dec{-1.18} & 79.70 \inc{+0.70} & 1153.86 \dec{-17.50} & 651.79 \inc{+106.07} & 57.40 \dec{-0.30} & 71.01 \inc{+0.00} & 69.50 \dec{-0.41} & 51.80 \dec{-0.30} \\
CoT & 41.76 \dec{-3.58} & 78.90 \dec{-0.10} & 1507.67 \inc{+336.30} & 661.07 \inc{+115.36} & 57.50 \dec{-0.20} & 70.75 \dec{-0.26} & 66.23 \dec{-3.58} & 53.50 \inc{+1.40}\\
CCoT & 44.50 \dec{-0.84} & 79.00 \inc{+0.00} & 1289.67 \inc{+118.30} & 604.29 \inc{+58.58} & 53.00 \dec{-4.70} & 71.69 \inc{+0.68} & 69.70 \dec{-0.11} & 51.20 \dec{-0.90}\\
CoT + RED & 46.07 \inc{+0.73} & \textbf{81.00 \inc{+2.00}} & \textbf{1574.52 \inc{+403.15}} & \textbf{695.36 \inc{+149.65}} & \textbf{59.50 \inc{+1.80}} & 72.15 \inc{+1.14} & 70.73 \inc{+0.92} & \textbf{54.80 \inc{+2.70}}\\
CCoT + RED & \textbf{47.50 \inc{+2.16}} & 80.60 \inc{+1.60} & 1359.28 \inc{+187.91} & 642.50 \inc{+96.79} & 58.40 \inc{+0.70} & \textbf{72.76 \inc{+1.75}} & \textbf{71.09 \inc{+1.28}} & 53.50 \inc{+1.40}\\
\midrule
\multicolumn{9}{@{}l}{\cellcolor{LightYellow}\textbf{Qwen2.5-VL-7B}} \\
Baseline & 60.88 & 82.10 & 1665.22 & 621.07 & 56.70 & 58.13 & 77.76 & 64.70 \\
VCD & 59.34 \dec{-1.54} & 82.20 \inc{+0.10} & 1650.80 \dec{-34.42} & 631.07 \inc{+10.00} & 58.30 \dec{-0.40} & 60.32 \inc{+2.19} & 77.53 \dec{-0.23} & 64.50 \dec{-0.20}\\
VCD + ICD & 59.40 \dec{-1.48} & 82.40 \inc{+0.30} & 1584.17 \dec{-101.04} & 651.79 \inc{+30.71} & 57.50 \dec{-1.20} & 60.71 \inc{+2.58} & 75.45 \dec{-2.31} & 64.80\inc{+0.10}\\
CoT & 46.70 \dec{-14.18} & 81.80 \dec{-0.30} & 1555.52 \dec{-109.70} & 705.00 \inc{+83.93} & 56.80 \inc{+0.10} & 60.49 \inc{+2.36} & 70.32 \dec{-7.44} & 61.30 \dec{-3.40} \\
CCoT & 46.69 \dec{-14.19} & 81.00 \dec{-1.10} & 1559.71 \dec{-105.51} & 634.29 \inc{+13.22} & 52.90 \dec{-3.80} & 73.18 \inc{+15.05} & 74.60 \dec{-3.16} & 61.30 \dec{-3.40} \\
CoT + RED & 61.06 \inc{+0.18} & 82.20 \inc{+0.10} & \textbf{1706.25 \inc{+41.03}} & \textbf{706.79 \inc{+86.72}} & \textbf{60.70 \inc{+4.00}} & 76.50 \inc{+18.37} & 77.98 \inc{+0.22} & \textbf{70.60 \inc{+5.90}} \\
CCoT + RED & \textbf{61.92 \inc{+1.04}} & \textbf{84.60 \inc{+2.50}} & 1704.83 \inc{+39.61} & 648.21 \inc{+27.14} & 56.80 \inc{+0.10} & \textbf{78.37 \inc{+20.24}} & \textbf{78.61 \inc{+0.85}} & 68.10 \inc{+3.40} \\
\midrule
\multicolumn{9}{@{}l}{\cellcolor{LightRed}\textbf{Llama3-LLaVA-Next-8B}} \\
Baseline & 65.22 & 65.60 & \textbf{1583.67} & 332.14 & 37.70 & 72.57 & 65.01 & 37.70 \\
VCD & 63.60 \dec{-1.62} & 67.60 \inc{+2.00} & 1512.10 \dec{-71.57} & 341.79 \inc{+9.64} & 40.50 \inc{+2.80} & 71.93 \dec{-0.64} & 63.51 \dec{-1.50} & 35.90 \dec{-1.80}\\
VCD + ICD & 63.88 \dec{-1.34} & 70.50 \inc{+4.90} & 1485.12 \dec{-98.55} & 317.50 \dec{+14.64} & 40.30 \inc{+2.60} & 71.71 \dec{-0.86} & 62.97 \dec{-2.03} & 36.20 \dec{-1.50}\\
CoT & 61.82 \dec{-3.40} & 65.60 \inc{+0.00} & 1462.63 \dec{-121.04} & 415.71 \inc{+83.57} & \textbf{42.10 \inc{+4.40}} & 72.38 \dec{-0.19} & 64.05 \dec{-0.96} & 36.50 \dec{-1.20}\\
CCoT & 60.43 \dec{-4.79} & 63.50 \dec{-2.10} & 1424.86 \dec{-158.81} & 343.57 \inc{+11.43} & 35.80 \dec{-1.90} & 72.74 \inc{+0.17} & 64.18 \dec{-0.83} & 36.70 \dec{-1.00}\\
CoT + RED & 65.48 \inc{+0.26} & 74.10 \inc{+8.50} & 1583.56 \dec{-0.11} & 410.36 \inc{+78.22} & 40.60 \inc{+2.90} & 73.19 \inc{+0.62} & \textbf{66.66 \inc{+1.65}} & \textbf{39.10 \inc{+1.40}}\\
CCoT + RED & \textbf{65.91 \inc{+0.69}} & \textbf{76.50 \inc{+10.90}} & 1562.72 \dec{-20.95} & \textbf{441.79 \inc{+109.65}} & \textbf{42.10 \inc{+4.40}} & \textbf{73.22 \inc{+0.65}} & 66.04 \inc{+1.03} & 38.80 \inc{+1.10}\\
\bottomrule
\end{tabular}
}
\end{table*}

\begin{table*}[t]
\centering
\caption{\textbf{Detailed performance analysis on SEED-I for different CoT strategies}. 
RED's outputs evolve in accordance with the characteristics of specified CoT strategies.
For example, CoT with natural text rationales enhances text understanding, while CCoT with structured scene graph rationales improves visual reasoning capability.
}
\label{tab:detailed_seed_benchmark}
\setlength{\tabcolsep}{4pt} 
\resizebox{\linewidth}{!}{
\begin{tabular}{@{}lccccccccc@{}}
\toprule
 & \rotatebox[origin=l]{60}{\makecell[l]{\textbf{Scene}\\\textbf{Understanding}}}
 & \rotatebox[origin=l]{60}{\makecell[l]{\textbf{Instance}\\\textbf{Identity}}}
 & \rotatebox[origin=l]{60}{\makecell[l]{\textbf{Instance}\\\textbf{Attributes}}}
 & \rotatebox[origin=l]{60}{\makecell[l]{\textbf{Instance}\\\textbf{Location}}}
 & \rotatebox[origin=l]{60}{\makecell[l]{\textbf{Instances}\\\textbf{Counting}}}
 & \rotatebox[origin=l]{60}{\makecell[l]{\textbf{Spatial}\\\textbf{Relation}}}
 & \rotatebox[origin=l]{60}{\makecell[l]{\textbf{Instance}\\\textbf{Interaction}}}
 & \rotatebox[origin=l]{60}{\makecell[l]{\textbf{Visual}\\\textbf{Reasoning}}}
 & \rotatebox[origin=l]{60}{\makecell[l]{\textbf{Text}\\\textbf{Understanding}}} \\
\midrule
\multicolumn{10}{@{}l}{\cellcolor{LightBlue}\textbf{Gemma-3-4B}} \\
Baseline & 75.14 & 70.18 & 69.22 & 55.11 & 56.63 & 44.14 & \textbf{72.16} & 76.74 & 45.88 \\
CoT & 72.99 \dec{-2.15} & 70.02 \dec{-0.16} & 68.51 \dec{-0.71} & 57.41 \inc{+2.30} & 55.21 \dec{-1.42} & 49.92 \inc{+5.78} & 65.98 \dec{-6.18} & 72.21 \dec{-4.53} & 51.76 \inc{+5.88} \\
CCoT & 74.89 \dec{-2.25} & \textbf{70.84 \inc{+0.66}} & 70.23 \inc{+1.01} & \textbf{57.87 \inc{+2.76}} & 54.68 \dec{-1.95} & 44.81 \inc{+0.67} & 67.01 \dec{-5.15} & 75.23 \dec{-1.51} & 43.53 \dec{-2.35} \\
CoT + RED & 75.14 \inc{+0.00} & \textbf{70.84 \inc{+0.66}} & 69.89 \inc{+0.67} & 57.46 \inc{+2.35} & 55.21 \dec{-1.42}  & \textbf{50.68 \inc{+6.54}} & 67.01 \dec{-5.15} & 74.62 \dec{-2.12} & \textbf{52.94 \inc{+7.06}} \\
CCoT + RED & \textbf{75.93 \inc{+0.79}} & 70.19 \inc{+0.01} & \textbf{71.52 \inc{+2.30}} & \textbf{57.87 \inc{+2.76}} & \textbf{56.86 \inc{+0.23}} & 45.66 \inc{+1.52} & 67.01 \dec{-5.15} & \textbf{77.34 \inc{+0.60}} & 43.53 \dec{-2.35} \\
\midrule
\multicolumn{10}{@{}l}{\cellcolor{Blue}\textbf{Gemma-4-12B}} \\
Baseline & 77.45 & 75.15 & 72.60 & 66.56 & 62.08 & 57.53 & 74.23 & 77.95 & 37.65 \\
CoT & 76.25 \dec{-1.20} & 74.11 \dec{-1.04} & 72.51 \dec{-0.09} & 63.80 \dec{-2.76} & 62.61 \inc{+0.53} & 61.04 \inc{+3.51} & 73.20 \dec{-1.03} & 76.74 \dec{-1.21} & 61.18 \inc{+23.53} \\
CCoT & 77.58 \inc{+0.13} & 75.42 \inc{+0.27} & 72.75 \inc{+0.15} & 65.85 \dec{-0.71} & \textbf{64.98 \inc{+2.90}} & 56.62 \dec{-0.91} & 74.23 \inc{+0.00} & \textbf{79.15 \inc{+1.20}} & 60.00 \inc{+22.35} \\
CoT + RED & 77.64 \inc{+0.19} & 75.20 \inc{+0.05} & 74.25 \inc{+1.65} & 64.93 \dec{-1.63} & 63.38 \inc{+1.30} & \textbf{63.01 \inc{+5.48}} & \textbf{76.29 \inc{+2.06}} & 78.55 \inc{+0.60} & \textbf{63.53 \inc{+25.88}} \\
CCoT + RED & \textbf{78.28 \inc{+0.83}} & \textbf{75.97 \inc{+0.82}} & \textbf{74.79 \inc{+2.19}} & \textbf{68.10 \inc{+1.54}} & 64.53 \inc{+2.45} & 59.82 \inc{+2.29} & 75.26 \inc{+1.03} & \textbf{79.15 \inc{+1.20}} & 60.59 \inc{+22.94} \\
\midrule
\multicolumn{10}{@{}l}{\cellcolor{LightYellow}\textbf{Qwen2.5-VL-7B}} \\
Baseline & 66.37 & 61.11 & 68.85 & 40.18 & 32.04 & 49.16 & 72.16 & 74.92 & 45.88 \\
CoT & 69.06 \inc{+2.69} & 62.37 \inc{+1.26} & 71.26 \inc{+2.41} & 45.19 \inc{+5.01} & 33.76 \inc{+1.72} & 51.29 \inc{+2.13} & 73.20 \inc{+1.04} & 77.04 \inc{+2.12} & 50.59 \inc{+4.71} \\
CCoT & 73.34 \inc{+6.97} & 65.21 \inc{+4.10} & 74.70 \inc{+5.85} & 54.29 \inc{+14.11} & 38.08 \inc{+6.04} & 54.49 \inc{+5.33} & 74.23 \inc{+2.07} & 77.95 \inc{+3.03} & 54.12 \inc{+8.24} \\
CoT + RED & 80.34 \inc{+13.97} & 81.21 \inc{+20.10} & \textbf{81.33 \inc{+12.48}} & 72.70 \inc{+32.52} & \textbf{73.80 \inc{+41.76}} & \textbf{66.97 \inc{+17.81}} & 74.23 \inc{+2.07} & 80.66 \inc{+5.74} & \textbf{84.71 \inc{+38.83}} \\
CCoT + RED & \textbf{80.37 \inc{+14.00}} & \textbf{81.70 \inc{+20.59}} & 81.05 \inc{+12.20} & \textbf{73.11 \inc{+32.93}} & 73.64 \inc{+41.60} & 64.99 \inc{+15.83} & \textbf{76.29 \inc{+4.13}} & \textbf{81.27 \inc{+6.35}} & 77.65 \inc{+31.77} \\
\midrule
\multicolumn{10}{@{}l}{\cellcolor{LightRed}\textbf{Llama3-LLaVA-Next-8B}} \\
Baseline & 77.77 & 76.52 & 76.23 & 65.24 & 64.36 & 52.21 & \textbf{75.26} & 76.13 & 55.29 \\
CoT & 77.23 \dec{-0.54} & 75.42 \dec{-1.10} & 77.16 \inc{+0.93} & 66.97 \inc{+1.73} & 61.10 \dec{-3.26} & \textbf{55.71 \inc{+3.50}} & 72.16 \dec{-3.10} & 73.41 \dec{-2.72} & 56.65 \inc{+1.36}\\
CCoT & 76.88 \dec{-0.89} & 76.95 \inc{+0.43} & 77.35 \inc{+1.12} & \textbf{67.08 \inc{+1.84}} & 63.10 \dec{-1.26} & 54.34 \inc{+2.13} & 70.10 \dec{-5.16} & 75.83 \dec{-0.30} & 51.76 \dec{-3.53} \\
CoT + RED & \textbf{78.75 \inc{+0.98}} & \textbf{77.06 \inc{+0.54}} & \textbf{77.69 \inc{+1.46}} & 66.67 \inc{+1.43} & 64.12 \dec{-0.24} & 54.79 \inc{+2.58} & 73.20 \dec{-2.06} & \textbf{78.25 \inc{+2.12}} & \textbf{64.71 \inc{+9.42}} \\
CCoT + RED & 78.28 \inc{+0.51} & \textbf{77.50 \inc{+0.98}} & \textbf{78.55 \inc{+2.32}} & \textbf{67.08 \inc{+1.84}} & \textbf{64.94 \inc{+0.58}} & 53.27 \inc{+1.06} & 74.23 \dec{-1.03} & 77.04 \inc{+0.91} & 57.65 \inc{+2.36} \\
\bottomrule
\end{tabular}
}
\end{table*}

\subsection{Settings}\label{sec:setting}
\textbf{Baselines.}
The baselines include: \uline{Baseline} (standard inference with $p_\theta(y_i|\bm{y}_{<i},x,q)$), \uline{CoT}~\cite{Wei_NeurIPS22_CoT,Zhang_TMLR_mm-cot} (prompting to generate text rationales), and \uline{CCoT}~\cite{Mitra_CVPR24_CCoT} (a state-of-the-art training-free method generating JSON-formatted scene graph rationales).
Other plug-and-play decoding baselines are:
\uline{VCD}~\cite{Leng_CVPR24_VCD}, contrasting $p_\theta(y_i|\bm{y}_{<i},x,q)$ with $p_\theta(y_i|\bm{y}_{<i},x^\prime,q)$, where $x^\prime$ is a corrupted input image by adding Gaussian noise,
and \uline{VCD + ICD}~\cite{Wang_ACL24F_ICD}, improving VCD using a variant query $q^\prime$ modified by adversarial instruction.
While they were proposed for object hallucination in LVLMs, comparing RED with them helps validate the practicality because they reportedly improved general performance~\cite{Leng_CVPR24_VCD,Wang_ACL24F_ICD}.\looseness-1

\textbf{Benchmark Datasets.}
We used six diverse visual reasoning benchmark datasets:  \uline{GQA}~\cite{Hudson_CVPR19_gqa}, \uline{TextVQA}~\cite{Singh_CVPR19_textvqa}, \uline{MME}~\cite{Fu_2023_mme}, \uline{SEED-I}~\cite{Li_CVPR24_seed_bench}, \uline{LLaVA-Bench}~\cite{Liu_CVPR24_improved_llava_llavabench}, \uline{MM-Vet}~\cite{Yu_ICML24_mmvet}, and \uline{MathVista}~\cite{Lu_ICLR24_mathvista}.
These benchmarks are generally used for evaluating various types of multi-modal capabilities, including general question answering, visual recognition, OCR, mathematical reasoning, multi-modal comprehension, and spatial understanding.

\textbf{Models.}
We used publicly available LVLMs on HuggingFace~\cite{Wolf_arXiv19_huggingface}: Gemma-3 (4B, 12B, 27B)~\cite{Google_2025_gemma3}, Qwen-2.5-VL (7B, 32B, 72B)~\cite{Bai_2025_Qwen-vl}, and Llama3-LLaVA-Next (8B)~\cite{Li_2024_llavanext}.

\textbf{Evaluation Protocols.}
We used greedy decoding.
$\lambda$ for RED was chosen from \{0.1, 0.3, 0.5, 1.0, 10.0\} using scores on validation or development sets; We discuss the effect of $\lambda$ in Appendix.
Tables show scores with parenthesized relative \textcolor{blue}{improvement} or \textcolor{red}{degradation} from Baseline.\looseness-1

\subsection{Evaluation on General Visual Reasoning Tasks}
Table~\ref{tab:general_benchmark} shows RED's general visual reasoning improvements.
While CoT/CCoT sometimes outperformed Baseline, they were inconsistent, with large drops on some LVLM-benchmark pairs, especially TextVQA, which requires understanding texts in images.
In contrast, RED consistently improved CoT/CCoT and outperformed all baselines in nearly all cases, even on TextVQA.
RED thus successfully addresses the issues of the existing multi-modal CoT using $p_\theta(y_i|\bm{y}_{<i},x,r,q)$ in terms of performance.
The consistent improvements offered by RED highlight the potential value of intermediate rationales generated in multi-modal CoT and its practicality for leveraging in diverse domains.

\subsection{Detailed Analysis for Multi-modal Capabilities}
Table~\ref{tab:detailed_seed_benchmark} summarizes the detailed performance analysis with respect to nine multi-modal capabilities of LVLMs provided by the SEED-I benchmark.
Notably, RED amplified multi-modal capabilities depending on the nature of the given rationale, i.e., detailed textual descriptions in CoT and JSON-formatted scene graphs in CCoT.
On the one hand, CoT + RED significantly enhanced text understanding and spatial relation capabilities, which require detailed information to be accurate.
On the other hand, CCoT + RED remarkably improved instance attributes and instance localization, which can be clearly described using semi-structured data such as JSON.
These observations indicate that RED can appropriately condition the rationale content to the final output and assign different capabilities to LVLMs according to the characteristics of a given rational format.

\subsection{Intervention Analysis}\label{sec:intervention}
We examine whether RED can overcome the issue of multi-modal CoT not being able to leverage rationale, as described in Section~\ref{sec:motivation}.
We hypothesized that the performance should improve with higher-quality rationales and degrade with irrelevant ones.
To evaluate this, we conduct intervention analysis using GPT-4~\cite{Achiam_2023_GPT4} generated rationales as the high-quality rationale, and the randomly swapped irrelevant rationales in the same way as Section~\ref{sec:motivation}.
Table~\ref{tab:intervention} shows the results of this intervention analysis on GQA with CCoT, where \textbf{Self} denotes using the rationales generated from the backbone LVLM itself, \textbf{GPT-4} denotes using high-quality rationales generated by GPT-4, and \textbf{Random} denotes using irrelevant rationales swapped from other instances.
Note that we used the same $\lambda$ tuned with the Self rationales across all cases.
CCoT + RED with GPT-4 rationales greatly improved performance, Random rationales degraded it.
The substantial performance gains observed with CCoT + RED using GPT-4 rationales, exceeding the CCoT counterparts, indicate that RED effectively leverages improved rationale content for multi-modal reasoning.
These results indicate not only that RED can overcome the issue of conventional multi-modal CoT reasoning but also that RED offers some interpretability in the reasoning process, which is dependent on the rationale.

\begin{table}[t]
\centering
\centering
\resizebox{\linewidth}{!}{
\begin{tabular}{@{}lccc@{}}
\toprule
 & \textbf{Self} ($\uparrow$) & \textbf{GPT-4} ($\uparrow$) & \textbf{Random} ($\downarrow$) \\
\midrule
\multicolumn{4}{@{}l}{\cellcolor{LightBlue}\textbf{Gemma-3-4B}} \\
CCoT      & 44.54 \inc{+4.54} & 45.79 \inc{+5.79} & 44.35 \inc{+4.35} \\
CCoT + RED  & \textbf{45.87 \inc{+5.87}} & \textbf{48.93 \inc{+8.93}} & 39.36 \dec{-0.64} \\
\midrule
\multicolumn{4}{@{}l}{\cellcolor{Blue}\textbf{Gemma-12-4B}} \\
CCoT      & 44.50 \dec{-0.84} & 45.61 \inc{+0.27} & 44.30 \dec{-1.04} \\
CCoT + RED  & \textbf{47.50 \inc{+2.16}} & \textbf{50.04 \inc{+4.70}} & 43.29 \dec{-2.05} \\
\midrule
\multicolumn{4}{@{}l}{\cellcolor{LightYellow}\textbf{Qwen2.5-VL-7B}} \\ 
CCoT      & 46.69 \dec{-14.19} & 50.85 \dec{-10.03} & 51.76 \dec{-9.12} \\
CCoT + RED  & \textbf{61.92 \inc{+1.04}}  & \textbf{63.11 \inc{+2.23}}  & 41.74 \dec{-19.14} \\
\midrule
\multicolumn{4}{@{}l}{\cellcolor{LightRed}\textbf{Llama3-LLaVA-Next-8B}} \\ 
CCoT      & 60.43 \dec{-4.79} & 57.84 \dec{-7.38} & 60.91 \dec{-4.31} \\
CCoT + RED  & \textbf{65.91 \inc{+0.69}} & \textbf{67.88 \inc{+2.66}} & 58.70 \dec{-6.52} \\
\bottomrule
\end{tabular}
}
\caption{\textbf{Intervention analysis on GQA with CCoT}. Performance is shown when using self-generated rationales (Self), rationales generated by GPT-4 (GPT-4), and randomly swapped irrelevant rationales (Random).
RED can be improved by a higher-quality rationale and degraded by a lower-quality one, demonstrating its faithfulness in grounding on rationales.
}
\label{tab:intervention} 
\end{table}
\begin{figure}[t]
  \centering
  \includegraphics[width=\linewidth]{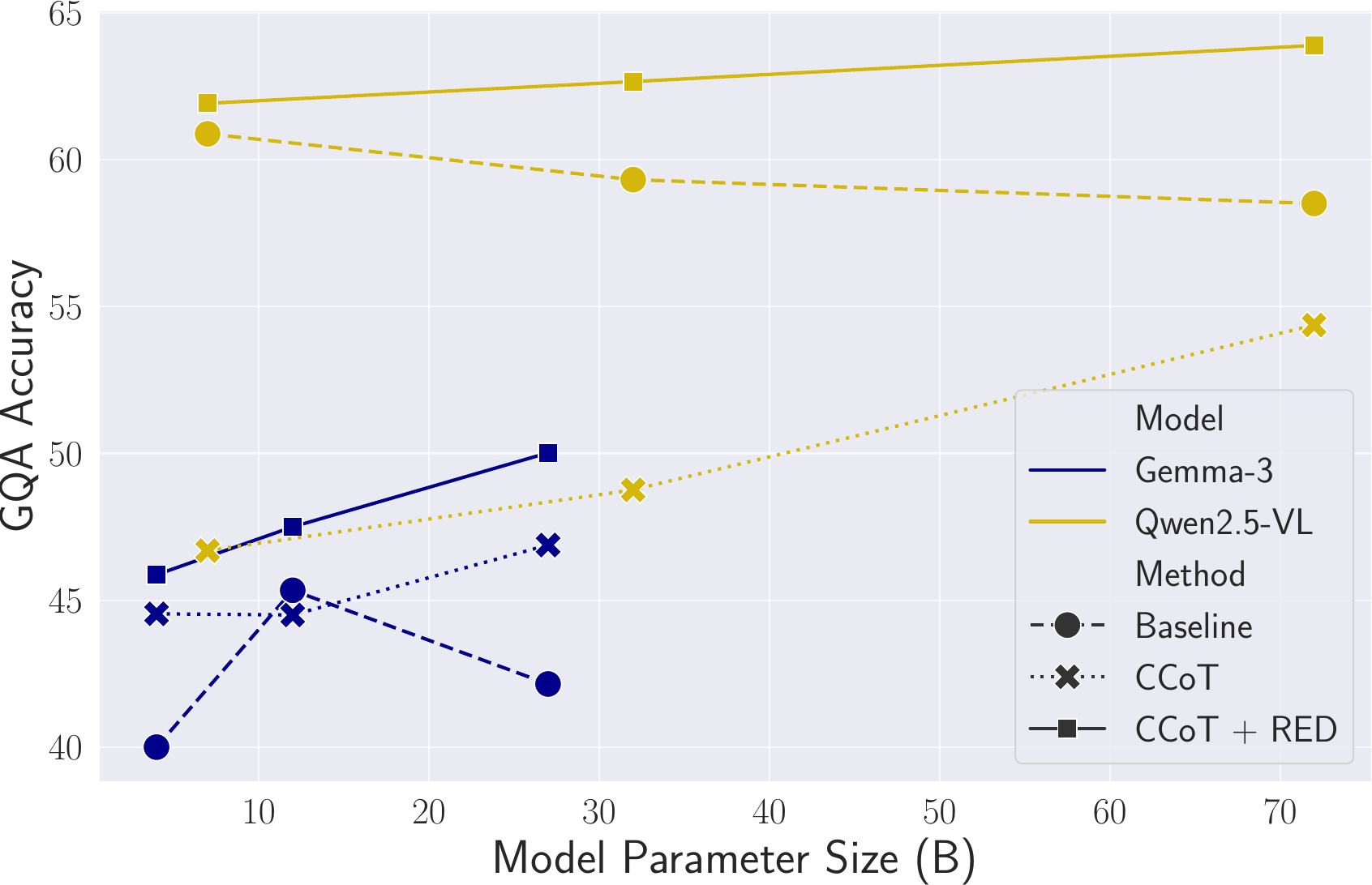}
  \captionof{figure}{
  \textbf{Performance trends on GQA accuracy with increasing LVLM parameter sizes} (Baseline, CCoT, CCoT + RED for Gemma-3 and Qwen2.5-VL families).
  RED can consistently improve Baseline and CCoT in any models, and can unlock further performance scalability according to model sizes.
  }
  \label{fig:scalability}
\vspace{-3mm}
\end{figure}

\subsection{Scalability for Larger LVLMs}
We test RED on LVLMs with a larger parameter size for evaluating the scalability.
We additionally utilized Gemma-3-27B, Qwen2.5-VL-32B/-72B.
These are increased language model sizes, and the vision encoders remain unchanged.
Figure~\ref{fig:scalability} shows the scalability on larger LVLMs.
Contrary to intuition, neither Baseline nor CCoT consistently improved performance with model size.
This is potentially because, as shown in Section~\ref{sec:motivation}, the decoding process often over-relies on visual input tokens at the expense of rationales; a similar observation is found in~\cite{Al-tahan_NeurIPS24_unibench}.
In contrast, our RED consistently improved performance in proportion to model size, indicating that RED can leverage sophisticated rationales from larger language models.
This implies that RED could be a key factor in unlocking the full potential of larger LVLMs by enabling more effective multi-modal CoT.

\subsection{Qualitative Evaluation}
\begin{figure}[t] 
  \centering
      \includegraphics[width=\linewidth]{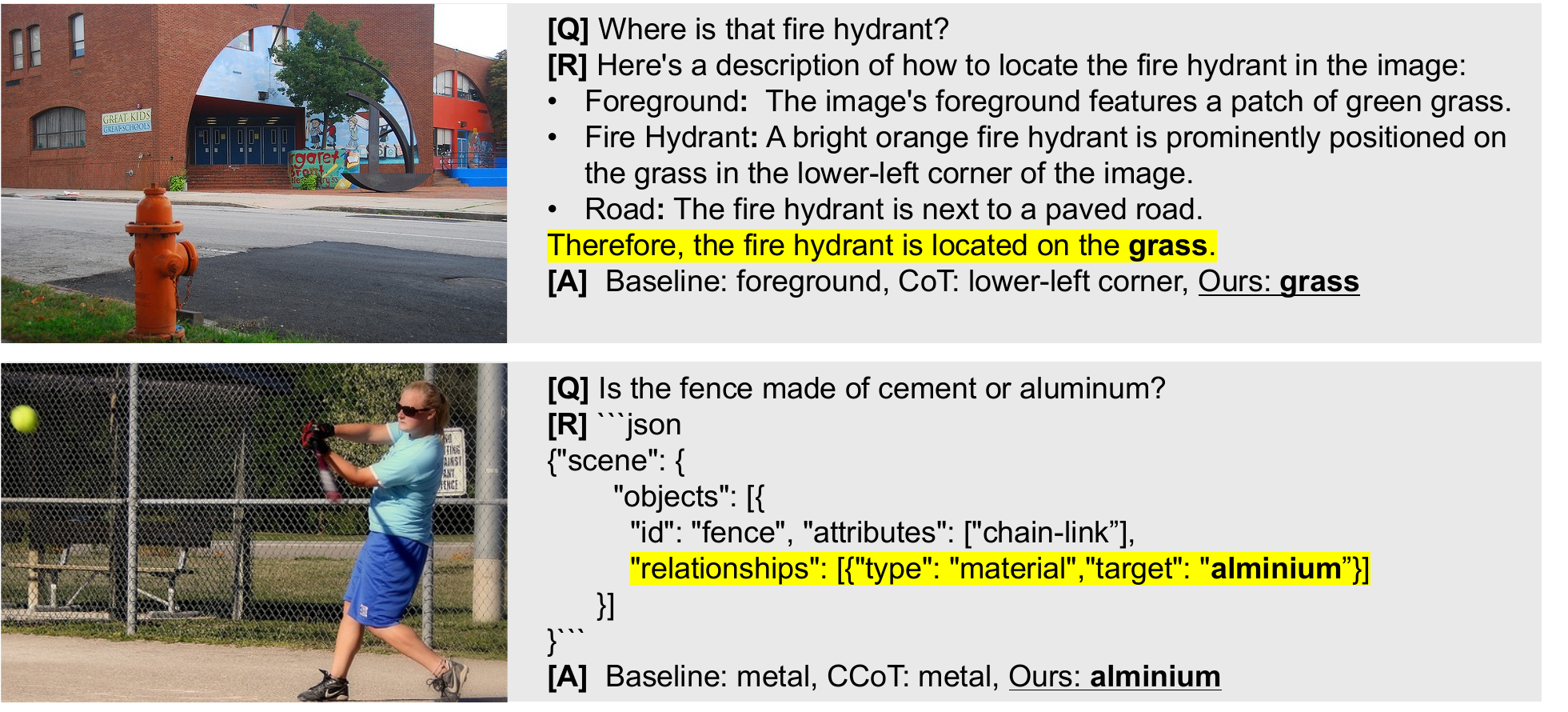}
      \captionof{figure}{
      \textbf{Qualitative examples of CoT reasoning on GQA (Gemma-3-12B)}. For each example, the input image (left), the query [Q], the generated rationale [R] (text for CoT and JSON for CCoT), and the answers from Baseline, CoT/CCoT, and our method (RED) [A] are shown. RED successfully leverages the rationale to produce the correct answer.}
      \label{fig:visualization}
  \vspace{-3mm}
\end{figure}

Figure~\ref{fig:visualization} shows qualitative examples from GQA; we performed reasoning with CoT and CCoT at the top and bottom of the figure, respectively.
While generated rationales point to the correct answer, naive CoT/CCoT produce incorrect responses, highlighting their failure to leverage the rationales by $p_\theta(y_i|\bm{y}_{<i},x,r,q)$.
In contrast, RED extracts the conclusion and related attributes from the rationales in both cases.\looseness-1


\subsection{Inference Efficiency}
We evaluate the inference efficiency of RED.
The inference steps are decomposed into (i) rationale generation by Eq.~\eqref{eq:rationale_generation} and (ii) answer generation by Eq.~\eqref{eq:red}. 
The former is the same as the existing CoT, and the latter requires a dual-forward pass, which we implemented using batch-parallel decoding. 
To assess the latency, we measured the average inference time for each query across all benchmarks with Gemma-3-12B on Baseline, CoT, and RED.
The results were 3.01, 5.27, and 5.05 for Baseline, CoT, and RED, respectively.
RED was faster than CoT because the contexts of $(x,q)$ and $(r,q)$ of the batch-parallel decoding were shorter than the full context $(x,r,q)$ of CoT, emphasizing the efficiency of our method.
Nonetheless, slightly increasing GPU memory consumption due to parallel RED is a limitation that should be resolved in future work.

\looseness-1

\section{Conclusion}\label{sec:conclusion}
This paper addressed the challenge of LVLMs ineffectively using rationales in CoT reasoning. 
We proposed Rationale-Enhanced Decoding (RED), an optimal, plug-and-play solution derived from KL-constrained reward maximization, which harmonizes visual and rationale inputs by multiplying their distinct conditional distributions.
Extensive experiments demonstrated RED's consistent and significant reasoning improvements over existing methods.

A notable limitation is the increased inference overhead, a common trade-off for plug-and-play decoding strategies. Future efforts could focus on mitigating this cost. Despite this, RED offers a significant step towards more reliable, interpretable, and rationale-grounded multi-modal systems.

\clearpage

{
    \small
    \bibliographystyle{ieeenat_fullname}
    \bibliography{ref}
}

\clearpage
\appendix

\section{Setting of Preliminary Experiments in Section 2.3}\label{append:preexp_setting}
All experiments were done with greedy decoding on Gemma-3-12B. We used the test set of GQA for the evaluations.

For computing attention contribution scores, we used the outputs from the 15th transformer block, and averaged the scores across output tokens except for the last \texttt{EOS} token.
We calculated the attention contribution scores for each token type and displayed the average percentage between test samples in Figure~1.

For intervention analysis with an irrelevant rationale $r^\prime$, we generated $r^\prime$ by the same prompting as CoT/CCoT with the original tuple $(x,q)$.
Irrelevant image and query pair $(x^\prime,q^\prime)$ were uniform randomly selected from the other pairs in the test set of GQA.

\section{Additional Experiments}\label{append:red_ablation}
\subsection{Ablation Study of Eq.~(8)}
As alternatives for Eq.~(8), we used (i) rationale-conditional probability $p_\theta(y_i|\bm{y}_{<i},r,q)$, (ii) mixture-of-experts $(1-\lambda) p_\theta(y_i|\bm{y}_{<i},x,q)+\lambda p_\theta(y_i|\bm{y}_{<i},r,q)$, and (iii) reversed power-of-experts $p_\theta(y_i|\bm{y}_{<i},r,q)\times p_\theta(y_i|\bm{y}_{<i},x,q)^\lambda$.
We see that our formulation by Eq.~(8) achieves the best accuracy.
While the rationale-conditional probability $p_\theta(y_i|\bm{y}_{<i},r,q)$ is grounded on the rationale $r$ in the next token prediction, it may cause the over-reliance on textual information, as reported in~\cite{Wang_ACL24F_ICD}.
For mixture-of-experts, it produces the next tokens by the sum of $p_\theta(y_i|\bm{y}_{<i},x,q)$ and $p_\theta(y_i|\bm{y}_{<i},r,q)$, which may fail to effectively collaborate image- and rationale-conditional prediction because it prioritizes tokens that have a high probability assigned to one of the distributions, similar to an OR operation.
The result of reverse power-of-experts demonstrates the importance of maximizing the rationale-conditional probability as a reward in Eq.~(7).
Since the effects of the image $x$ in $p_\theta(y_i|\bm{y}_{<i},x,r,q)$ is already dominant as discussed in Section~2.3, predicting by $p_\theta(y_i|\bm{y}_{<i},r,q)\times p_\theta(y_i|\bm{y}_{<i},x,q)^\lambda$ further weaken the effects of the rationale $r$, leading a suboptimal performance similar to Baseline and CCoT.
In contrast, our formulation of RED can appropriately focus on tokens important for both the image- and rationale-conditional probabilities by the multiplication, like an AND operation.
More importantly, Eq.~(8) has the theoretical guarantee that maximizes the rationale-conditional probability while maintaining the image-conditional prediction capability, achieving more stable and better accuracy gain for multi-modal CoT prompting.

\begin{table}[htbp]
\centering
\caption{Ablation study varying next token probability formulation in Eq.~(8) with CCoT on GQA.}
\label{tab:hal_benchmark}
\resizebox{\linewidth}{!}{
\begin{tabular}{@{}llc}
\toprule
 & Formulation & GQA Accuracy (\%)\\
\midrule
\multicolumn{3}{@{}l}{\cellcolor{Blue}\textbf{Gemma-3-12B}} \\
Baseline & $p_\theta(y_i|\bm{y}_{<i},x,q)$ & 45.34 \\
CCoT & $p_\theta(y_i|\bm{y}_{<i},x,r,q)$ & 44.50 \dec{-0.84} \\
Rationale-Conditional & $p_\theta(y_i|\bm{y}_{<i},r,q)$ & 37.87 \dec{-7.47} \\
VCD + ICD + CCoT & $ (1+\lambda) p_\theta(y_i|\bm{y}_{<i},x,r,q) - \lambda p_\theta(y_i|\bm{y}_{<i},x^\prime,r,q^\prime)$ & 44.09 \dec{-1.25} \\
Mixture-of-Experts & $(1-\lambda) p_\theta(y_i|\bm{y}_{<i},x,q)+\lambda p_\theta(y_i|\bm{y}_{<i},r,q)$ & 44.67 \dec{-0.67}\\
Rev. Power-of-Experts & $p_\theta(y_i|\bm{y}_{<i},r,q)\times p_\theta(y_i|\bm{y}_{<i},x,q)^\lambda$ & 44.01 \dec{-1.33} \\
Power-of-Experts (Eq.~(\ref{eq:red})) & $p_\theta(y_i|\bm{y}_{<i},x,q)\times p_\theta(y_i|\bm{y}_{<i},r,q)^\lambda$ & \textbf{47.50 \inc{+2.16}}\\
\bottomrule
\end{tabular}
}
\end{table}

\subsection{Effects on Object Hallucination}
We evaluate RED on two object hallucination benchmarks for LVLMs: MMHal~\cite{Sun_ACL24F_mmhal} and POPE~\cite{Li_EMNLP23_pope}.
Table~\ref{tab:hal_benchmark} shows the results with multiple LVLM backbones.
While RED is not particularly designed to mitigate object hallucination in LVLMs, it improved CoT/CCoT and achieved competitive performance with the specialized decoding methods for object hallucination, i.e., VCD and VCD + ICD.
In contrast to these specialized methods, RED boosts not only anti-hallucination capability but also the quality of the response, as shown in MMHal's average scores.
The improvements tend to increase when using high-performance LVLMs (e.g., Qwen2.5-VL-7B), implying the importance of rationale in RED as discussed in Section~4.4.

\begin{table*}[htbp]
\centering
\caption{Hallucination Benchmark}
\label{tab:hal_benchmark}
\begin{minipage}[]{0.47\linewidth}
\centering
\resizebox{\linewidth}{!}{
\begin{tabular}{@{}lccc}
\toprule
 & \multicolumn{2}{c}{\textbf{MMHal}} & \multirow{2}{*}{\textbf{POPE (All)}} \\
 \cmidrule(r){2-3}
 &  \textbf{Avg. Score ($\uparrow$)} & \textbf{Hal. Ratio} ($\downarrow$) &\\
\midrule
\multicolumn{4}{@{}l}{\cellcolor{LightBlue}\textbf{Gemma-3-4B}} \\
Baseline & 2.73 & 0.61 & 83.20 \\
VCD & 2.73 \inc{+0.00} & 0.56 \inc{-0.05} & 83.23 \inc{+0.03} \\
VCD + ICD & 2.75 \inc{+0.02} & 0.59 \inc{-0.02} & 83.27 \inc{+0.07} \\
CoT & 2.23 \dec{-0.50} & 0.65 \dec{+0.04} & 83.37 \inc{+0.17} \\
CCoT & 2.08 \dec{-0.65} & 0.66 \dec{+0.05} & 83.33 \inc{+0.13} \\
CoT + RED & 2.73 \inc{+0.00} & 0.58 \inc{-0.03} & 83.53 \inc{+0.33} \\
CCoT + RED & 2.76 \inc{+0.03} & 0.54 \inc{-0.07} & 83.57 \inc{+0.37}\\
\midrule
\multicolumn{4}{@{}l}{\cellcolor{Blue}\textbf{Gemma-3-12B}} \\
Baseline & 3.10 & 0.52 & 84.07 \\
VCD & 3.21 \inc{+0.11} & 0.52 \inc{-0.00} & 84.67 \inc{+0.60} \\
VCD + ICD & 3.29 \inc{+0.19} & 0.51 \inc{-0.01} & 84.73 \inc{+0.66} \\
CoT & 2.65 \dec{-0.45} & 0.52 \inc{+0.00} & 84.17 \inc{+0.10} \\
CCoT & 3.01 \dec{-0.09} & 0.48 \inc{-0.04} & 84.60 \inc{+0.53}\\
CoT + RED & 3.36 \inc{+0.26} & 0.46 \inc{-0.06} & 84.43 \inc{+0.36}\\
CCoT + RED & 3.23 \inc{+0.13} & 0.45 \inc{-0.07} & 85.20 \inc{+1.13}\\
\bottomrule
\end{tabular}
}
\end{minipage}
\hfill
\begin{minipage}[]{0.47\linewidth}
\resizebox{\linewidth}{!}{
\begin{tabular}{@{}lccc}
\toprule
 & \multicolumn{2}{c}{\textbf{MMHal}} & \multirow{2}{*}{\textbf{POPE (All)}} \\
 \cmidrule(r){2-3}
 &  \textbf{Avg. Score ($\uparrow$)} & \textbf{Hal. Ratio} ($\downarrow$) &\\
\midrule
\multicolumn{4}{@{}l}{\cellcolor{LightYellow}\textbf{Qwen2.5-VL-7B}} \\
Baseline & 3.85 & 0.36 & 87.90 \\
VCD & 3.86 \inc{+0.01} & 0.34 \inc{-0.02} & 89.40 \inc{+1.50} \\
VCD + ICD & 3.85 \inc{+0.00} & 0.33 \inc{-0.03} & 89.90 \inc{+2.00} \\
CoT & 3.84 \dec{-0.01} & 0.35 \inc{-0.01} & 84.20 \dec{-3.70} \\
CCoT & 3.77 \dec{-0.08} & 0.35 \inc{-0.01} & 86.33 \dec{-1.57} \\
CoT + RED & 3.99 \inc{+0.14} & 0.33 \inc{-0.03} & 88.63 \inc{+0.73} \\
CCoT + RED & 3.93 \inc{+0.08} & 0.33 \inc{-0.03} & 88.47 \inc{+0.57} \\
\midrule
\multicolumn{4}{@{}l}{\cellcolor{LightRed}\textbf{Llama3-LLaVA-Next-8B}} \\
Baseline & 2.97 & 0.53 & 88.23 \\
VCD & 3.10 \inc{+0.13} & 0.47 \inc{-0.06} & 89.50 \inc{+1.27} \\
VCD + ICD & 3.12 \inc{+0.15} & 0.46 \inc{-0.07} & 89.43 \inc{+1.20} \\
CoT & 3.05 \inc{+0.08}  & 0.48 \inc{-0.05} & 87.00 \dec{-1.23} \\
CCoT & 3.15 \inc{+0.18} & 0.49 \inc{-0.04} & 86.83 \dec{-1.40} \\
CoT + RED & 3.16 \inc{+0.19} & 0.46 \inc{-0.07} & 88.86 \inc{+0.63} \\
CCoT + RED & 3.14 \inc{+0.17} & 0.47 \inc{-0.06} & 88.63 \inc{+0.40} \\
\bottomrule
\end{tabular}
}
\end{minipage}
\end{table*}

\subsection{Effects of $\lambda$}\label{append:lambda}
\begin{figure}[htpb] 
  \centering
      \includegraphics[width=\linewidth]{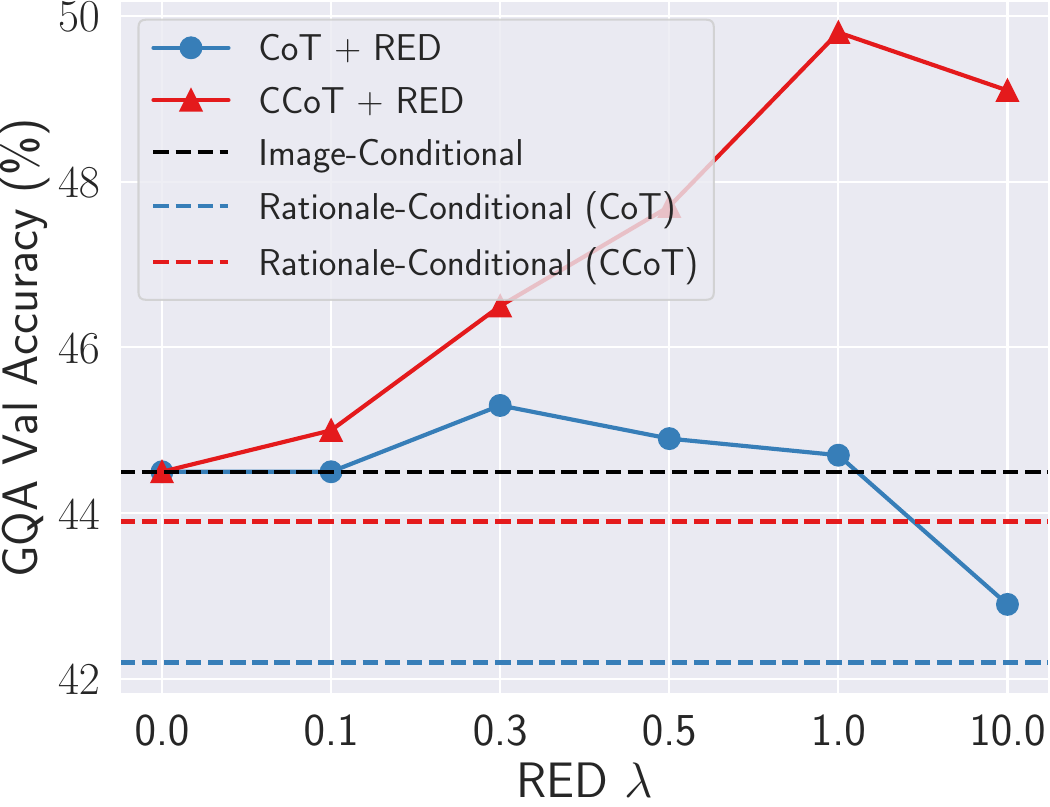}
      \captionof{figure}{Effects of $\lambda$ in RED on the GQA Validation Set.}
      \label{fig:lambda}
\end{figure}

We investigate the sensitivity of RED toward a selection of $\lambda$.
Figure~\ref{fig:lambda} shows the results on the validation set of GQA with Gemma-3-12B, where we vary $\lambda=\{0.0, 0.1, 0.3, 0.5, 1.0, 10.0\}$ and plot the baselines using image-conditional $p_\theta(y_i|\bm{y}_{<i},x,q)$ and rationale-conditional $p_\theta(y_i|\bm{y}_{<i},r,q)$.
We observe that the best value for $\lambda$ depends on the type of rationale, i.e., text description for CoT and JSON-formatted scene graph.
While the performance with larger $\lambda$ gradually approaches the performance of rationale-conditional $p_\theta(y_i|\bm{y}_{<i},r,q)$, RED stably outperform the image-conditional baseline unless choosing extremely large values.

\subsection{Evaluations on MMMU/MMMU-Pro}
Table~\ref{tab:mmmu} shows the results on the MMMU benchmarks.
RED largely outperformed baselines, indicating that it remains effective on more complex multi-modal reasoning tasks.

\begin{table}[t]
    \centering
    \caption{\footnotesize Evaluations on MMMU/MMMU-Pro with Qwen2.5-VL-7B}
    \label{tab:mmmu}
        \vspace{-2mm}
        \resizebox{0.9\columnwidth}{!}{
        \begin{tabular}{lcccc}
            \toprule
                  & \multicolumn{2}{c}{\uline{MMMU}} & \multicolumn{2}{c}{\uline{MMMU-Pro}}\\
            Model & Accuracy (\%) & Speed (ms/token) & Accuracy (\%) & Speed (ms/token) \\
            \midrule
            Baseline               & 50.3          & 9.5  & 35.2          & 12.4 \\
            VCD + ICD              & 48.5          & 12.1 & 33.1          & 14.7 \\
            CoT                    & 55.5          & 17.5 & 37.2          & 28.7 \\
            CoT + RED (Ours)       & \textbf{61.6} & 14.5 & \textbf{40.5} & 16.2 \\
            \bottomrule
        \end{tabular}
        }
        \vspace{-6mm}
\end{table}


\section{Broader Impacts}\label{sec:broader_impacts}
This work on rationale-enhanced decoding (RED) has the potential for several societal impacts, both positive and negative.

As a positive perspective, by making LVLMs' reasoning more grounded on rationales, RED can increase the transparency of these models. 
When users or developers can see that a model's output is a logical consequence of its intermediate reasoning steps (rationales), it can foster greater trust in AI systems.
This is particularly important for critical applications where understanding why an AI made a certain decision is essential.
Furthermore, when LVLMs are used to answer questions based on documents or to summarize information involving images and text, RED could ensure that the generated answers are more faithfully tied to the source material and the model's interpretation process, reducing the spread of misinformation stemming from model errors.

From a negative perspective, even if RED improves faithfulness, if the generated rationales themselves are flawed (due to biases in training data or limitations in the LVLM's rationale generation capability), users might still be unduly convinced by a seemingly logical but ultimately incorrect or biased output. 
This is a form of automation bias, where humans over-rely on AI-generated information, especially if it's presented with a ``reasoning chain.''
To overcome these potential issues, it is important to develop robust methods for detecting and mitigating biases in both rationale generation and the final outputs, even when they appear logically derived.


\end{document}